\documentclass[10pt,twocolumn,letterpaper]{article}

\usepackage{iccv}
\usepackage{times}
\usepackage{epsfig}
\usepackage{graphicx}
\usepackage{amsmath}
\usepackage{amssymb}

\usepackage{multirow}
\usepackage{subfigure}
\usepackage{tabularx}
\usepackage{color}
\usepackage{rotating}
\usepackage{diagbox}
\usepackage{bm}

\usepackage[pagebackref=true,breaklinks=true,colorlinks,bookmarks=false]{hyperref}

\iccvfinalcopy

\def\iccvPaperID{2697} 

\ificcvfinal\fi

\begin{document}

\title{Sparse Needlets for Lighting Estimation with Spherical Transport Loss}

\author{
Fangneng Zhan \textsuperscript{\rm 1,2,4},
Changgong Zhang \textsuperscript{\rm 2},
Wenbo Hu \textsuperscript{\rm 3},
Shijian Lu \textsuperscript{\rm 1}, \\ 
Feiying Ma \textsuperscript{\rm 2},
Xuansong Xie \textsuperscript{\rm 2}, 
Ling Shao \textsuperscript{\rm 4} \\
\textsuperscript{\rm 1} Nanyang Technological University \quad  \textsuperscript{\rm 2} DAMO Academy, Alibaba Group \\
\textsuperscript{\rm 3} The Chinese University of Hong Kong \quad  \textsuperscript{\rm 4} Inception Institute of Artificial Intelligence
}

\maketitle
\ificcvfinal\thispagestyle{empty}\fi

\begin{abstract}
Accurate lighting estimation is challenging yet critical to many computer vision and computer graphics tasks such as high-dynamic-range (HDR) relighting. Existing approaches model lighting in either frequency domain or spatial domain which is insufficient to represent the complex lighting conditions in scenes and tends to produce inaccurate estimation. This paper presents NeedleLight, a new lighting estimation model that represents illumination with needlets and allows lighting estimation in both frequency domain and spatial domain jointly. An optimal thresholding function is designed to achieve sparse needlets which trims redundant lighting parameters and demonstrates superior localization properties for illumination representation.
In addition, a novel spherical transport loss is designed based on optimal transport theory which guides to regress lighting representation parameters with consideration of the spatial information. Furthermore, we propose a new metric that is concise yet effective by directly evaluating the estimated illumination maps rather than rendered images.
Extensive experiments show that NeedleLight achieves superior lighting estimation consistently across multiple evaluation metrics as compared with state-of-the-art methods.
\end{abstract}

\begin{figure*}[t]
\centering
\subfigure [Garon et al. \cite{garon2019fast}]
{\includegraphics[width=.24\linewidth]{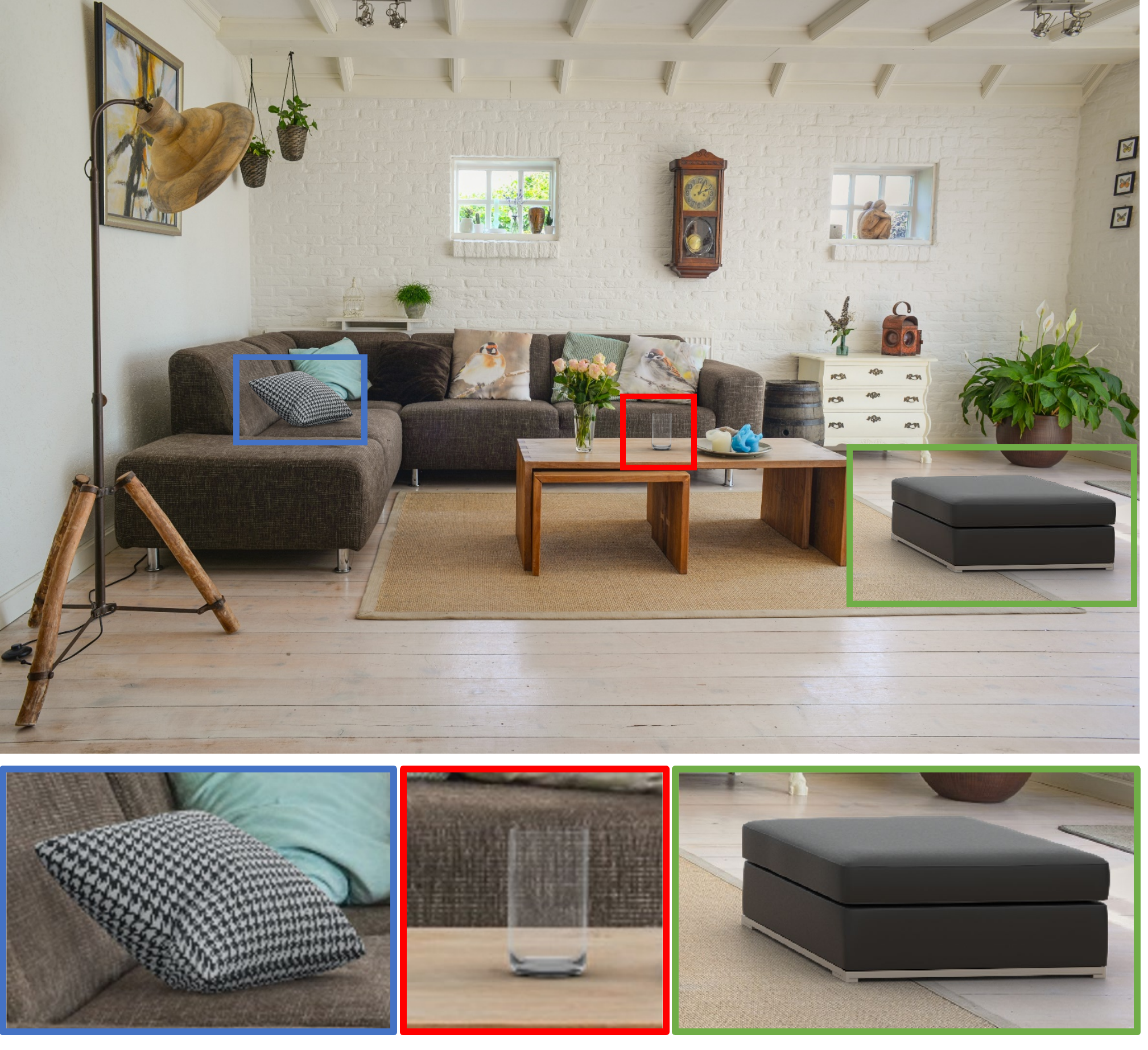}}
\subfigure [Gardner et al. \cite{gardner2019deep}] {\includegraphics[width=.24\linewidth]{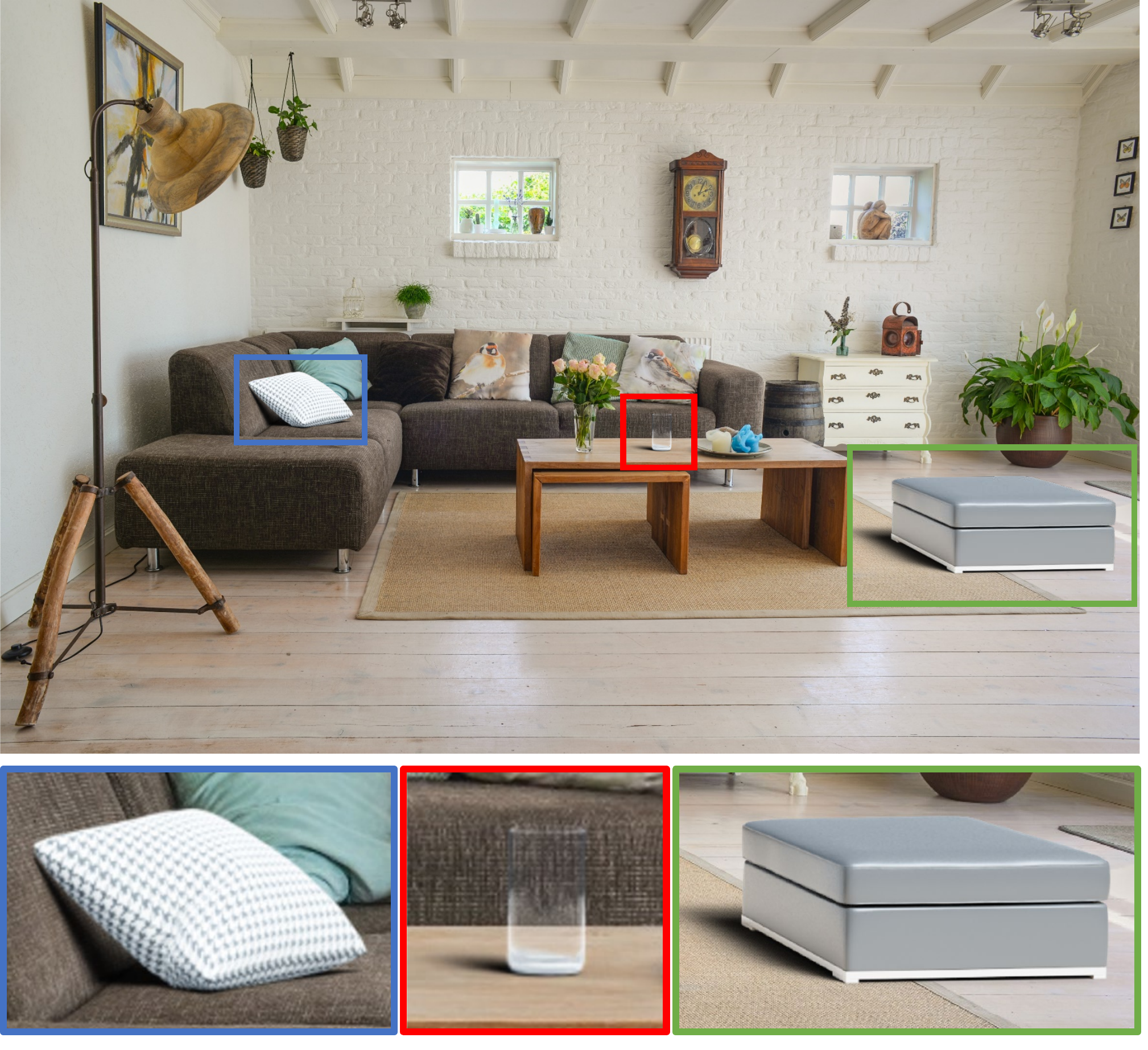}}
\subfigure [Ours] {\includegraphics[width=.24\linewidth]{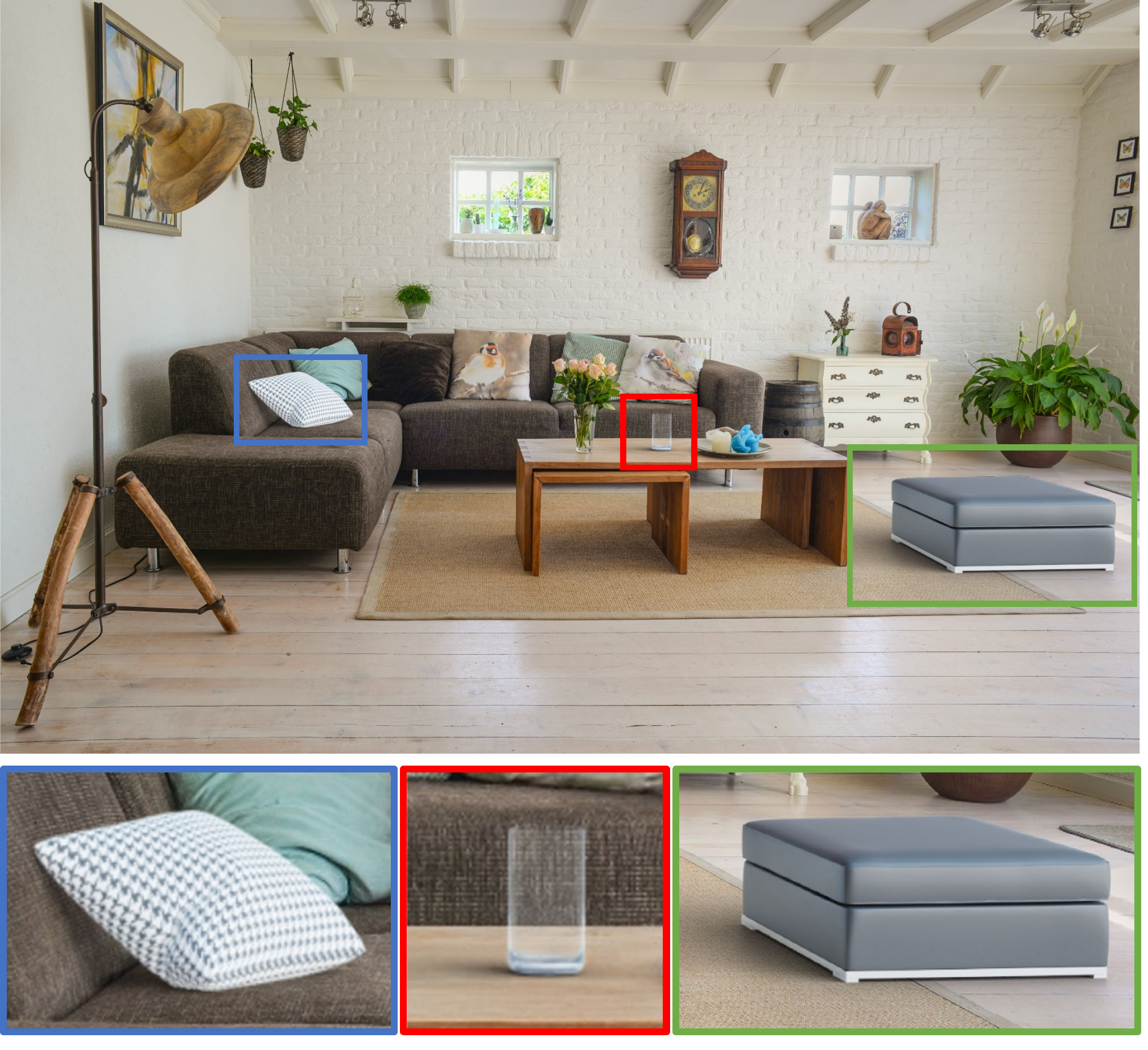}}
\subfigure [Ground Truth] {\includegraphics[width=.24\linewidth]{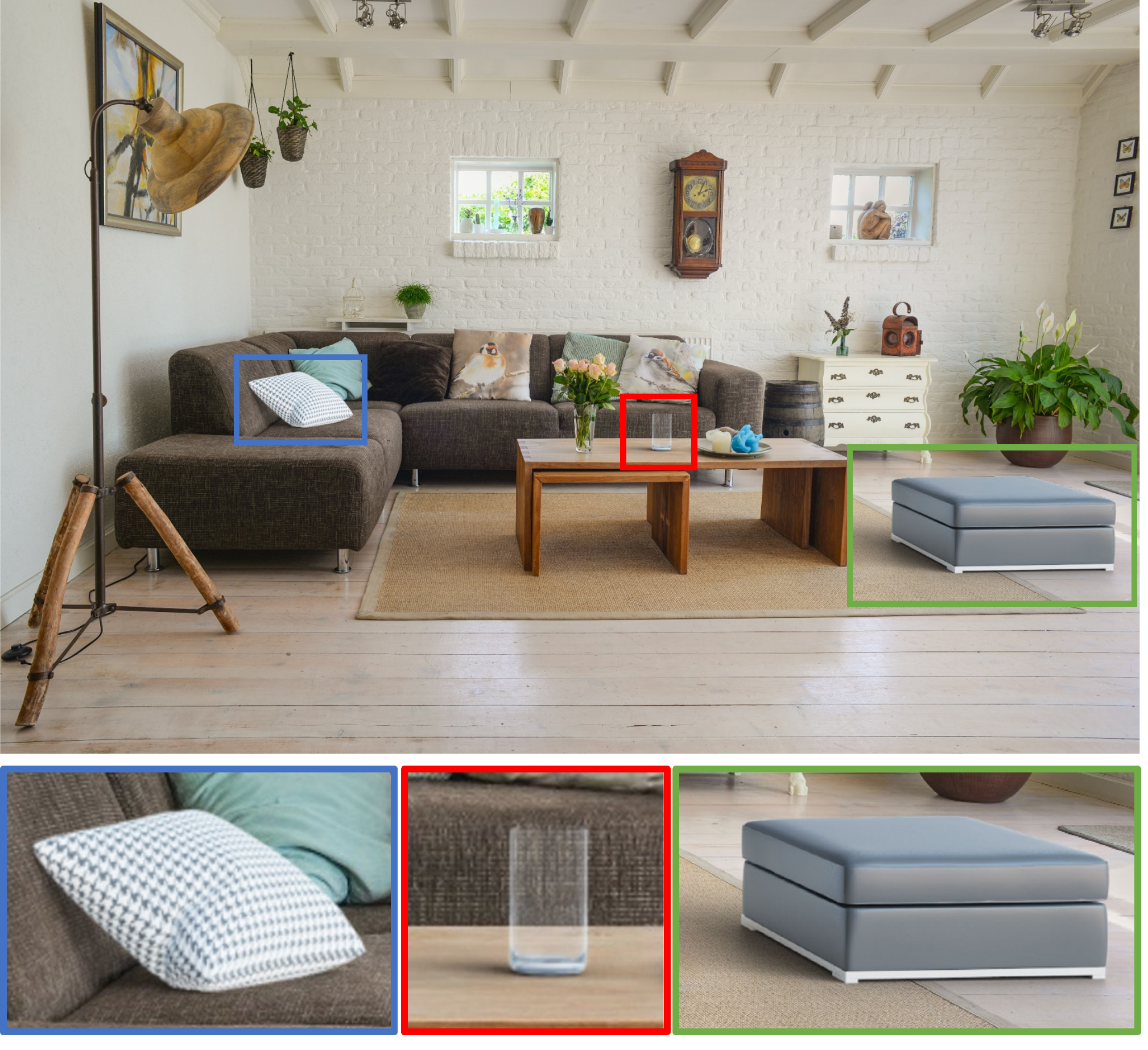}}
\caption{
The proposed NeedleLight estimates a parametric lighting representation from a single scene image which is critical to many tasks such as virtual object insertion.
Unlike previous methods that predict lighting in either frequency domain \cite{garon2019fast} (losing spatial localization) or spatial domain \cite{gardner2019deep} (losing frequency information) only, we introduce a novel needlets basis which is capable of representing and estimating lighting accurately in both frequency and spatial domains.
}
\label{im_intro}
\end{figure*}

\section{Introduction}
Lighting estimation aims to recover illumination from a single image with limited field of view. It has a wide range of applications in various computer vision and computer graphics tasks such as high-dynamic-range (HDR) relighting in mixed reality, etc. However, lighting estimation is a challenging and ill-posed problem as it needs to predict the illumination coming from a full sphere of directions including those unobserved from the current view in the scene. Additionally, it often requires to infer HDR illuminations from low-dynamic-range (LDR) observations so as to light virtual objects realistically while inserting them into real scene images as illustrated in Fig. \ref{im_intro}.

Lighting estimation has been tackled by regressing the parameters of various lighting representations in either frequency domain \cite{cheng2018shlight,garon2019fast}) or spatial domain \cite{gardner2017learning,gardner2019deep,song2019neural,srinivasan2020lighthouse}. However, lighting estimation in frequency domain normally represents illumination with Spherical Harmonics (SH) which lack spatial localization capabilities. Thus it tends to capture global lighting instead of the exact spatial locations of the light sources which often leads to weak shading and shadow effects as illustrated in Garon et al. \cite{garon2019fast} of Fig. \ref{im_intro}.
Lighting estimation in spatial domain has been addressed by direct generation of the illumination maps or indirect reconstruction through spherical Gaussian function. However, direct generation of illumination maps often leads to worse generalization as lighting estimation is an under-constrained problem by itself, and spherical Gaussian often involves a complicated training process as described in \cite{gardner2019deep}. Both 
types of approaches in spatial domain do not explicitly consider lighting frequency, and thus lead to inaccurate relighting performance as illustrated in Gardner et al. \cite{gardner2019deep} of Fig. \ref{im_intro}.
The high frequency information also tends to be blurred due to the use of naive L2 loss in the training.
Additionally, existing evaluation metrics in lighting estimation usually assess the objects rendered with the predicted illumination maps, which is time-consuming and sensitive to the test setting.

In this work, we propose NeedleLight, a new model that introduces needlet for accurate and robust lighting estimation from a single image. As a new generation of spherical wavelets, needlet enjoys good localization properties in both frequency and spatial domain which makes it ideal to be the basis for illumination representation.
Moreover, to remove the redundant parameters in needlet coefficients which will disturb the regression of principle light sources, we design an optimal thresholding function to achieve sparse needlets which improve the lighting estimation greatly.

Unlike spherical harmonic coefficients, needlet coefficients are spatially localized over a unit sphere.
To utilize the spatial information in regression, we propose a Spherical Transport Loss (STL) based on optimal transport theory.
STL is able to capture spatial information via a cost matrix and estimate the needlet coefficients more accurately than a naive L2 loss.
Besides, STL employs auxiliary point strategy to preserve high frequency information and greatly reduce the dimension of required parameters.
Based on STL, we design a new metric for the evaluation of lighting estimation by measuring the discrepancy between illumination maps.
The new metric highly simplifies the evaluation procedure and provides concise yet effective evaluation with regard to the lighting color, intensity and position.


The contribution of this work can be summarized in three aspects. 
First, we introduce a novel needlet basis for illumination representation which allows to regress the parameters in both frequency and spatial domains simultaneously. 
Second, we develop an optimal thresholding function to achieve sparse needlets which effectively removes the redundant needlet coefficients and improves the lighting estimation.
Third, we design a novel Spherical Transport Loss (STL) that effectively utilizes the spatial information of needlet coefficients in regression.
With STL, we also design a new evaluation metric that is more concise and effective than existing evaluation metrics.

\section{Related Works}
\textbf{Lighting Estimation:} Lighting estimation is a classic challenge in computer vision and computer graphics, and it is critical for realistic relighting in virtual objects insertion \cite{debevec2008rendering,zhan2018verisimilar,lalonde2012estimating,zoran2014shape,zhan2020towards,barron2013intrinsic,hold2017deep,zhan2021gmlight,murmann2019dataset,zhan2020aicnet,boss2020two,zhan2019scene} and image synthesis \cite{zhan2021unite,zhan2019esir,zhan2019gadan,zhan2019sfgan,xue2018accurate,zhan2021rabit,zhang2021deep,yu2021diverse,zhan2021spatial}. Traditional approaches require user intervention or assumptions about the underlying illumination model, scene geometry, etc. For example, Karsh et al. \cite{karsch2011rendering} recovers parametric 3D lighting from a single image but requires user annotations for initial lighting and geometry estimates. Zhang et al. \cite{zhang2016emptying} requires a full multi-view 3D reconstruction of scenes. Lombardi et al. \cite{lombardi2015reflectance} estimates illumination from an object of known shape with a low-dimensional model.

The recent works estimate lighting by regressing representation parameters or generating illumination maps \cite{li2020inverse,weber2018learning,hold2017deep,zhang2019all,liu2020lighting,chalmers2020reconstructing,park2020physically}.
For example, Cheng et al. \cite{cheng2018shlight} regresses the SH parameters of global lighting with a render loss. 
Maier et al. \cite{maier2017intrinsic3d} recover spherical harmonics illumination with additional depth information. 
Garon et al. \cite{garon2019fast} estimate lighting by predicting SH coefficients from a background image and local patch. 
Gardner et al. \cite{gardner2019deep} estimate the positions, intensities, and colours of light sources and reconstructs illumination maps with a spherical Gaussian function. 
Li et al. \cite{li2019deep} represent illumination maps with multiple spherical Gaussian functions and regresses the corresponding Gaussian parameters for lighting estimation. 
Gardner et al. \cite{gardner2017learning} generate illumination maps directly with a two-steps training strategy. 
Legendre et al. \cite{legendre2019deeplight} regress HDR lighting from LDR images by comparing the ground-truth sphere image to the rendered one with the predicted illumination. Srinivasan et al. \cite{srinivasan2020lighthouse} estimate a 3D volumetric RGB model of a scene and uses standard volume rendering to estimate incident illuminations.
Zhan et al. \cite{zhan2021emlight} propose to formulate the regression of illumination as the regression of a spherical distribution.

\textbf{Needlets:}
Needlets are a new generation of spherical wavelets \cite{wavelets} and have desirable localization capabilities in both spatial and frequency domains. Narcowich et al. \cite{narcowich2006needlets} first introduces needlets in the Functional Analysis literature, and Baldi et al. \cite{baldi2009asymptotics} further analyses the statistical properties of needlets. Due to the good localization property in the multipole and pixel spaces, needlets have been widely applied to the research of Cosmic Microwave Background \cite{pietrobon2006cmb,marinucci2008spherical}.

The aforementioned works estimate lighting in either frequency domain or spatial domain which is insufficient to capture the complex illumination in real scenes. Additionally, most existing works regress the illumination with a naive L2 loss or its variant which struggles to regress high frequency information and often introduces blurs. We introduce needlet basis for lighting representation which allows regression of illumination in both frequency and spatial domains jointly. A novel spherical transport loss is proposed to achieve stable and accurate regression of needlet coefficients. Details on lighting representation, spherical transport loss are to be presented in the ensuing sections.

\begin{figure*}[t]
\centering
\includegraphics[width=1.0\linewidth]{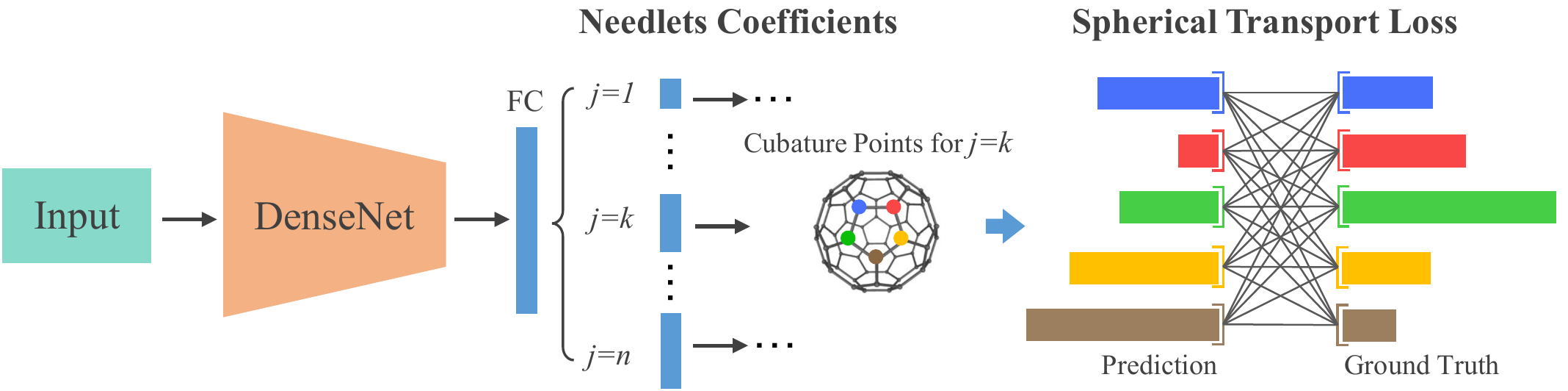}
\caption{The architecture of the proposed NeedleLight: We introduce sparse needlets to represent lighting.
The frequency bands in \textit{Needlet Coefficients} are denoted as $j=1,\cdots,n$. The needlet coefficients in each frequency band are spatially localized on a set of \textit{Cubature Points} on the unit sphere (illustration for $j=k$ only).
For stable lighting regression, we design a spherical transport loss 
to capture the divergence between the predicted needlet coefficients (\textit{Prediction}) and the ground-truth coefficients (\textit{Ground Truth}). The values of needlet coefficients are illustrated by color bars (only draw 5 for illustration).
}
\label{im_stru}
\end{figure*}

\begin{figure}[ht]
\centering
\includegraphics[width=1.0\linewidth]{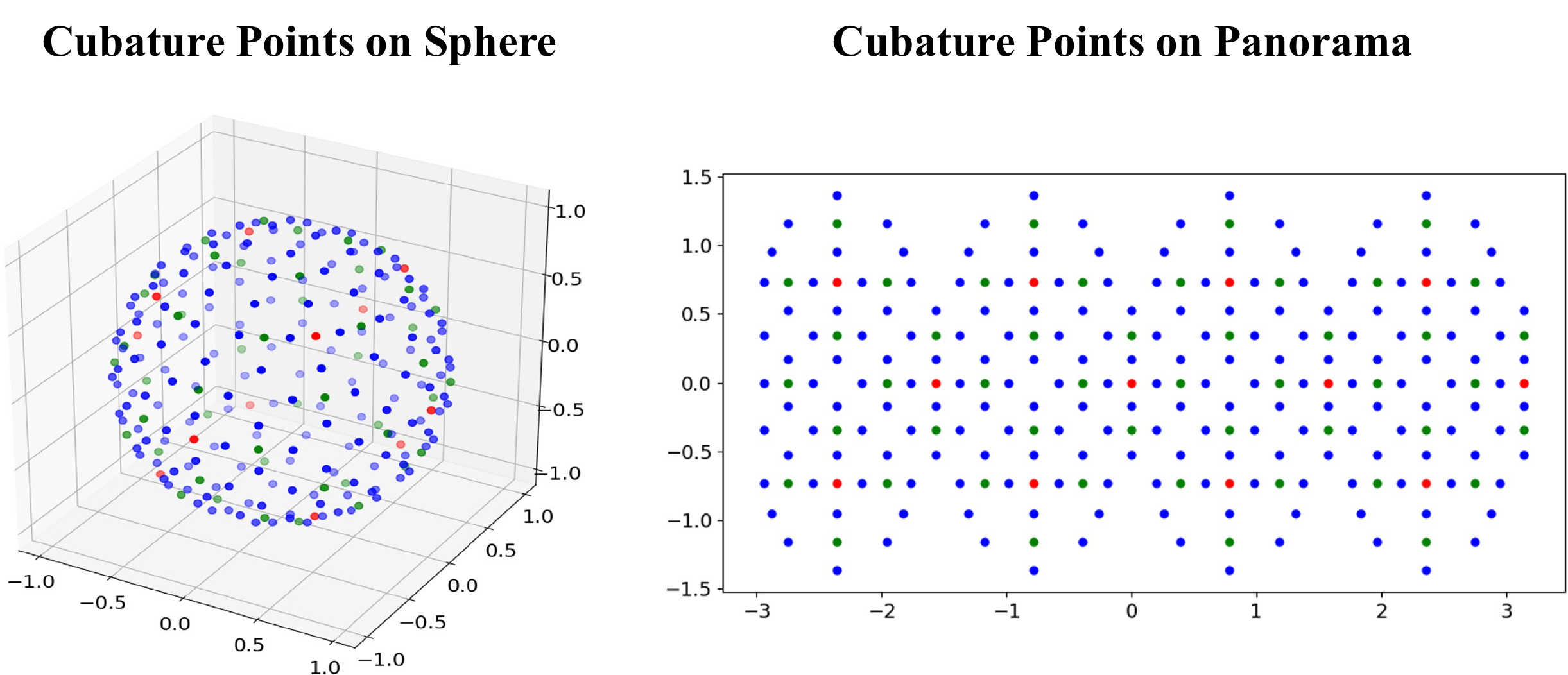}
\caption{
The visualization of cubature points ($j_{max}$=3) on sphere and panorama. The red, green and blue points denote the cubature points of frequency bands from low to high frequency, namely $j$=1, $j$=2 and $j$=3 respectively.
}
\label{im_cubature}
\end{figure}

\section{Proposed Method}

The proposed NeedleLight estimates illumination by regressing needlet coefficients from a single image as illustrated in Fig. \ref{im_stru}. A novel spherical transport loss is designed to achieve stable and effective regression of needlet coefficients. The following subsections describe how needlets and spherical transport loss work together for accurate and robust lighting estimation.

\subsection{Needlets based Lighting Representation}
Needlets \cite{ baldi2009asymptotics,narcowich2006needlets} are a new type of spherical wavelets that has been successfully used in microwave signal analysis. 
They can be localized at a finite number of frequencies, and decay quasi-exponentially fast away from the global maximum. Thus they enjoy good localization properties in both frequency and spatial domains. As described in \cite{baldi2009asymptotics}, a signal $I(x)$ (e.g. lighting signals of interest in this research) at a given frequency $j \in \mathbb{N}$ can be represented by the spherical needlets basis $\psi_{jk}(x)$ and the needlet coefficients $\beta_{jk}$ as follows:
\begin{small}
\begin{equation}
\begin{split}
    & \psi_{jk}(x) = \sqrt{\lambda_{jk}} \sum_{l=\lceil B^{j-1} \rceil} ^{\lfloor B^{j+1} \rfloor} b(\frac{l}{B^{j}}) \sum_{m=-l}^{l} Y_{lm}(\xi_{jk}) \overline{Y}_{lm}(x) \\
    & \beta_{jk} = \sqrt{\lambda_{jk}} \sum_{l=0}^{\infty}b(\frac{l}{B^{j}})\sum_{m=-l}^{l}a_{lm}Y_{lm}(\xi_{jk}) 
\end{split}
\end{equation}
\end{small}
where $x \in \mathbb{S}^{2}$, 
$\xi_{jk}$ and $\lambda_{jk}$ are pre-defined cubature points that spread over the unit sphere as shown in Fig. \ref{im_cubature}, and the associated cubature weights, respectively,
$b(\cdot)$ is a window function, $B$ is a free parameter larger than 1,
$Y_{lm}$ is spherical harmonic function with degree $l$ and order $m$, $a_{lm}$ is the corresponding spherical harmonic coefficients,
$\overline{Y}_{lm}$ is the complex conjugation of ${Y}_{lm}$.
$\xi_{jk}$ represents the spatial location of needlet basis $\psi_{jk}$, thus the needlet coefficients are spatially localized on the unit sphere. The signal $I(x)$ can be reconstructed via $I(x) = \sum_{j,k} \beta_{jk}\psi_{jk}(x)$.

Compared with SH, needlets have compact supports for the localization in spatial domain. As a result, they can easily and parsimoniously represent signals over the unit sphere that exhibits local sharp peaks or valleys, which are commonly presented in HDR illumination maps. Consequently, needlets serve as a more suitable basis for the representation of illumination maps.

\begin{figure*}[t]
\centering
\subfigure []
{\includegraphics[width=.28\linewidth]{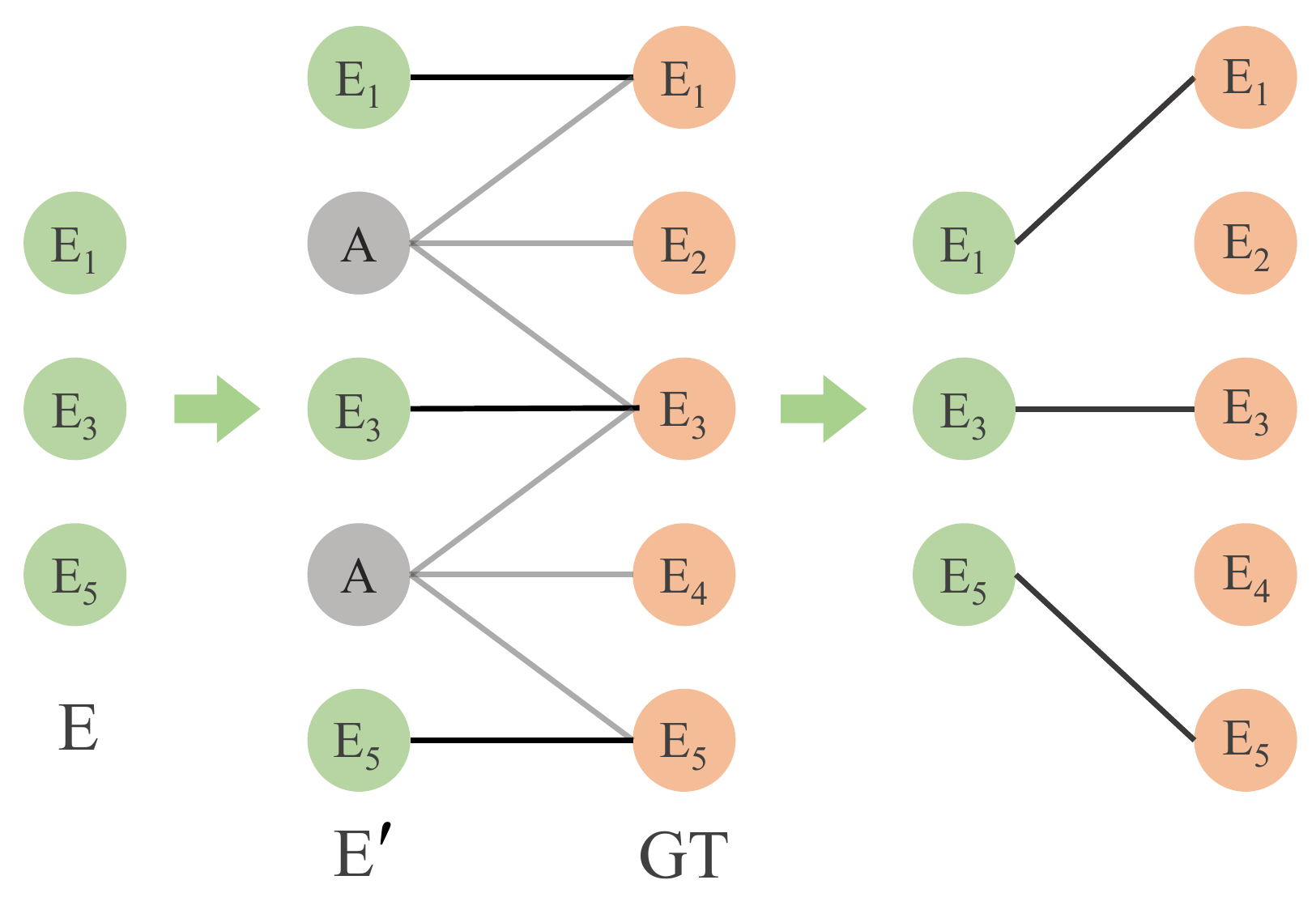}}
\hspace{10 pt}
\subfigure []
{\includegraphics[width=.36\linewidth]{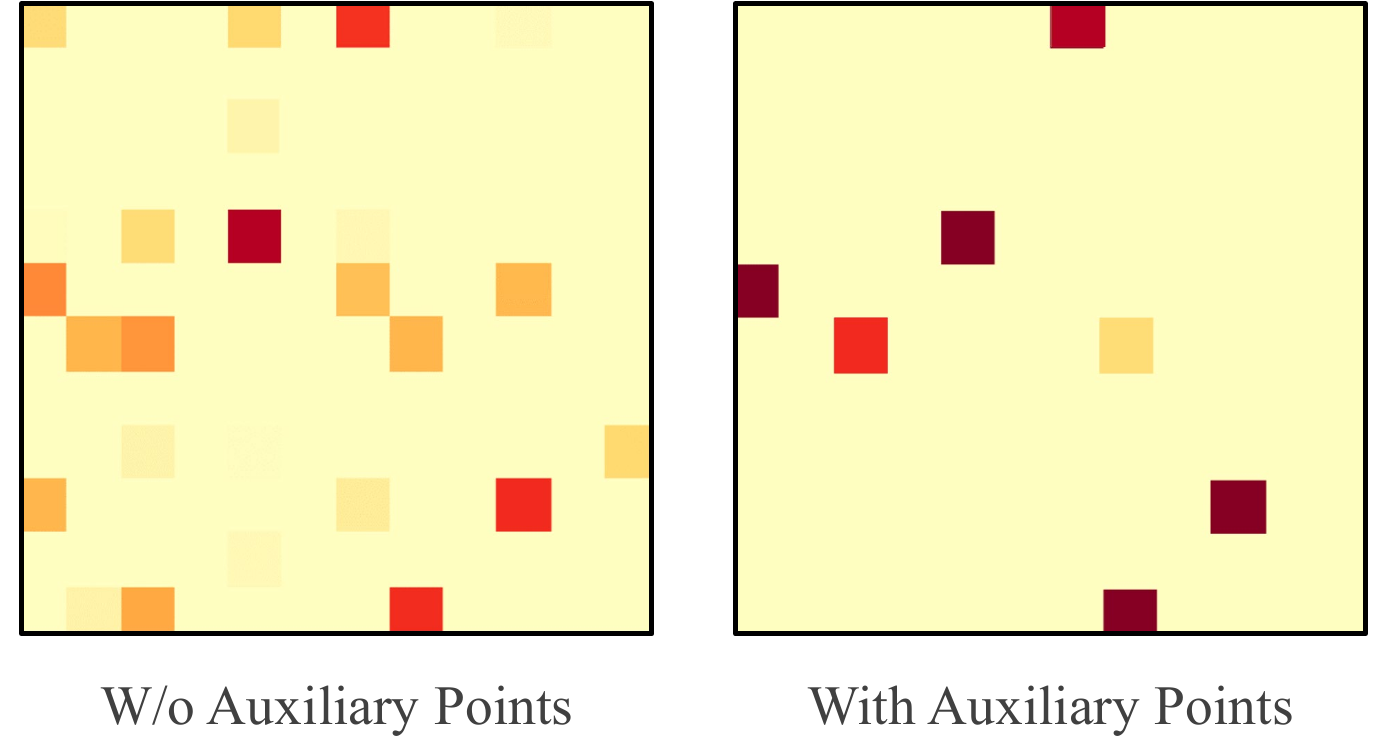}}
\hspace{10 pt}
\subfigure []
{\includegraphics[width=.26\linewidth]{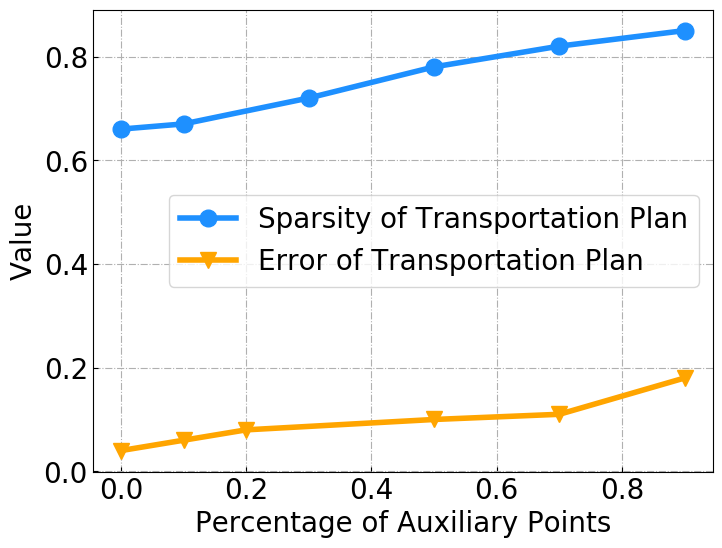}}
\caption{(a) The process to achieve sparse transportation plan through auxiliary points. E and GT are the estimated needlet coefficients and the ground truth, respectively. `A' denotes the auxiliary point. E$'$ denotes the predicted coefficients extended with auxiliary points. The connection between E$'$ and GT signifies the resulting transportation plan.
(b) Transportation plans for the needlet coefficients (j=1) w/o and w/ auxiliary points, respectively.
(c) The sparsity and error of transportation plan matrix with different percentage of auxiliary points.
}
\label{im_sparse}
\end{figure*}

\subsection{Sparse Needlets}
A signal is said to be sparse if it can be reconstructed with only a small amount of basis functions.
An illumination map consists of several dominant light sources with high radiance energy and an ambient residual. The light sources are obviously sparse under the needlet basis while the ambient component is not. For the lighting estimation, the reconstruction of dominant light sources is more significant than the reconstruction of the ambient component especially in high frequency sections.
Besides, there are dozens of needlet coefficients in high frequency section (252 coefficients for $j_{max}$=3), most of which are redundant parameters of ambient. Those redundant parameters will severely disturb the regression of principal light sources and lead to diffuse illumination (or low frequency illumination) which is undesired for relighting. Thus we deduce an optimal sparse function for needlets to separate the principle coefficients of light sources from the redundant ambient component.

We derive the sparse function from a Bayesian perspective and form the problem as a maximum posterior estimator.
Normally, we assume that the needlet coefficients of light sources $s$ follows Laplace distribution prior  as Laplace prior is well adapted to model sparse signals \cite{bobin2008sz}, which is also proved in \cite{dror2004statistical}. The needlet coefficients of ambient can be ideally treated as a Gaussian distribution \cite{maino2002all}.
The needlet coefficients $\beta$ of the illumination map can thus be modelled as:
\begin{equation}
    \beta = \underbrace{s}_{light \; sources} + \underbrace{\phi}_{ambient} + \underbrace{\eta}_{noise} 
\end{equation}
where $s$ denotes the needlet coefficients of sparse light sources which follow Laplace distribution,
$\phi$ denotes the needlet coefficients of ambient which is a Gaussian distribution, and $\eta$ denotes noises that follow a Gaussian distribution.
According to Vansynge et al. \cite{vansyngel2016semi}, the Bayesian formulation of the problem can be written as:
\begin{equation}
\begin{split}
& P(s | \beta) \propto P(s) \mathcal{L}(\beta|s) \\
& \mathcal{L}(\beta|s, \phi)  = N(\beta;\phi+s, M_{\eta})  \\
\end{split}
\end{equation}
where $M_{\eta}$ denotes the covariance matrices of the noise. As $\phi$ follows Gaussian distribution, we can derive:
\begin{footnotesize}
\begin{equation}
    \begin{split}
& \mathcal{L} (\beta| s) = \int ... \int \mathcal{L}(\beta|s, \phi) P(\phi)d\phi \propto   \\
& \exp \left[ -\frac{1}{2}\beta^{T}M_{\eta}^{-1}\beta + s^{T}M_{\eta}^{-1}\beta - \frac{1}{2}s^{T} M_{\eta}^{-1}s \right] \cdot \exp \\
& \left[ \frac{1}{2}(M_{\eta}^{-1}\beta - M_{\eta}^{-1}s)^{T}(M_{\eta}^{-1} + M_{\phi}^{-1})(M_{\eta}^{-1}\beta - M_{\eta}^{-1}s) \right]
    \end{split}
\end{equation}
\end{footnotesize}
where $M_{\phi}$ denotes the covariance matrices of the Gaussian distribution $\phi$. As $P(s)$ follows the Laplace distribution namely $P(s) \propto \exp \left[ -\lambda ||s|| \right]$, the maximum posterior estimator is obtained by
maximizing $P(s|\beta) = \mathcal{L}(\beta|s)*P(s)$. 
We take the partial derivative with respect to $s$:
\begin{footnotesize}
\begin{equation}
\begin{split}
    & \partial_{s}(-log(P(s|\beta))) = -(M_{\eta}+M_{\eta} M_{\phi}^{-1}M_{\eta})^{-1}s + 
    M_{\eta}^{-1}s \\
    & + (M_{\eta}+M_{\eta}M_{\phi}^{-1}M_{\eta})^{-1}\beta
    -M_{\eta}^{-1}\beta + \lambda \partial_{s} ||s||
\end{split}
\end{equation}
\end{footnotesize}
By making $\partial_{s}(-log(P(s|\beta))) = 0$, the following solution (or sparse function) can thus be derived:
\begin{equation}
    s = \beta - \left[M_{\eta}^{-1} - (M_{\eta}+M_{\eta} M_{\phi}^{-1}M_{\eta})^{-1} \right] ^{-1} \lambda \partial_{s} ||s|| 
\end{equation}
which is 
a soft thresholding operator with threshold $(M_{\eta}^{-1} - (M_{\eta}+ M_{\eta} M_{\phi}^{-1}M_{\eta})^{-1})\lambda$. More details about the derivation of the sparse function are provided in the supplemental file.

We apply the sparse function largely to the \textit{high-frequency needlet coefficients} so as to shrink the redundant coefficients of the ambient component. The sparse needlets provide sparse representation of illumination maps, which is desirable for the regression of light sources by using the proposed spherical transport loss to be described in the following subsection.

\subsection{Spherical Transport Loss}

Different from spherical harmonics, needlet coefficients are spatially localized in a unit sphere. A simple MSE loss cannot utilize the spatial information.
Besides, as an under-constrained task, the optimization in lighting estimation is severely challenging since it aims to recover the environment lighting from all directions based on a single scene image with a limited field of view. 
Thus the training of lighting estimation models struggles with the regression of high frequency information.

In this work, we propose a novel Spherical Transport Loss (STL) to achieve the stable and effective lighting regression. Because the needlet coefficients are spatially localized on the unit sphere as specified by cubature points, we treat the regression of needlet coefficients as an Unbalanced Optimal Transport (UOT) \cite{liero2018optimal,chizat2016scaling} problem on the unit sphere. Intuitively, UOT computes the cost for transporting a measure distributed on a space to another measure of possibly different total masses. 
As the needlet coefficients may contains negative values, we take the natural exponent of them before deriving the spherical transport loss.
Then we can define two sets of needlet coefficients represented by two positive vectors $a = (a_{1}, \cdot, a_{n}) \in \mathbb{R}^{n}_{+}$ and $b = (b_{1},\cdots,b_{n}) \in \mathbb{R}^n_{+}$ and their spatial layouts are specified by cubature points on the unit sphere. A distance matrix $C$ and a transportation plan matrix $P$ can thus be derived, where each entry $C_{ij}$ in $C$ gives the cost of moving point $a_{i}$ to point $b_{j}$ which can be defined by radian distance between points on the unit sphere, and $P_{ij}$ in $P$ represents the probability of assigning a point $a_{i}$ to a point $b_j$. The regularized UOT problem (namely STL) can thus be defined as follows:

\begin{footnotesize}
\begin{equation}
\begin{split}
& \mathop{min}\limits_{P} \left [ \langle C, P \rangle + \tau KL(P\cdot \vec{1} || a) + \tau KL (\vec{1} \cdot P || b) -  \gamma H(P) \right ]  \\
\end{split}
\end{equation}
\end{footnotesize}
where $\tau$ and $\gamma$ are regularization parameters, KL is the KerKullback-Leibler Divergence, $H(P)$ is the entropic regularization for efficient approximation of original UOT problem \cite{chizat2016scaling}.
The regularized UOT can be solved by Sinkhorn iteration \cite{cuturi2013sinkhorn}.

\begin{figure*}[ht]
\centering
\includegraphics[width=1.0\linewidth]{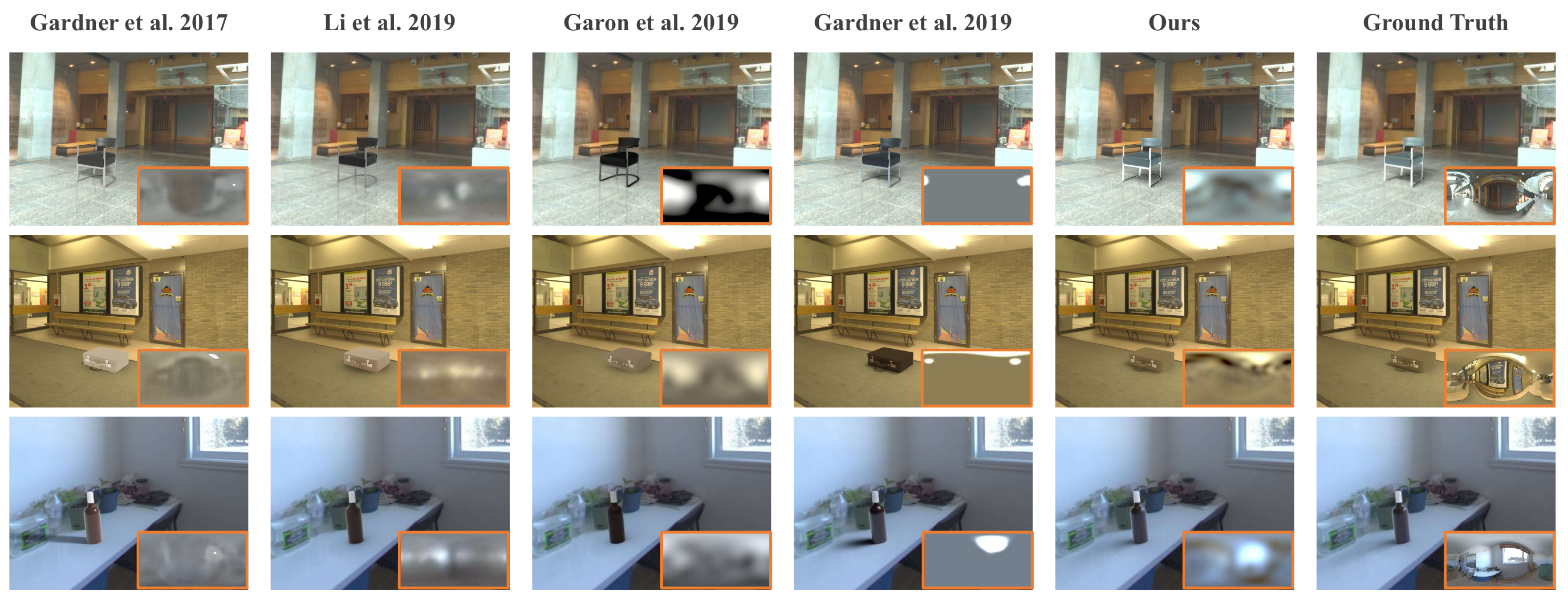}
\caption{Visual comparison of NeedleLight with state-of-the-art lighting estimation methods: With the illumination maps predicted by different methods (at bottom-right corner of each rendered image), the rendered virtual objects demonstrate different lighting intensity, color, shadow and shade.
}
\label{im_com}
\end{figure*}

To construct clear and sharp instead of diffuse (i.e., low frequency) light sources, we expect the transportation plan to be sparse. Otherwise the diffuse light sources will result in weak shading and shadow effects as presented in Garon et al. \cite{garon2019fast} of Fig. \ref{im_com}.
Besides, the dimension of output layer increases quickly when the frequency of needlets becomes large (252 needlet coefficients when $j_{max}=3$). We propose an auxiliary point strategy to achieve sparse transportation plan and reduce the dimension of output layer. Auxiliary points are assigned with small value (we select $\frac{1}{(number \; of \; coefficients)}$) and 0 cost for transport, which can be used for absorbing unused probability mass in cases of partial transport.
As shown in Fig. \ref{im_sparse}(a), we only estimate partial needlet coefficients (E), and then use the auxiliary points to replace other needlet coefficients to obtain a new set of coefficients (E$'$).
After obtaining the optimal transport between E$'$ and the ground truth GT, we can extract a sparse transportation plan by removing the connection with the auxiliary points. The two samples in Fig. \ref{im_sparse}(b) show the transportation plan matrix for the needlet coefficients ($j=1$) with and without auxiliary points, respectively.
Fig. \ref{im_sparse}(c) shows the sparsity and error of the transportation plan matrix ($n \times n$) with different percentage of auxiliary points
 (The evaluation metrics for the sparsity and error of transportation plan are described in \cite{blondel2018smooth}).
When more auxiliary points are used, the sparsity of the transportation matrix increases but the error also increases as shown in Fig. \ref{im_sparse} (c). As a trade-off, we select 66\% of coefficients as auxiliary points.

Using spherical transport loss for needlet coefficients regression has two clear advantages. First, it makes the regression sensitive to the global geometry, thus effectively penalizing predicted activation that is far away from the ground truth distribution. Second, it can preserve the high frequency information during training with the proposed auxiliary points.

\textbf{Evaluation Metric:}
Lighting estimation has been widely evaluated by using root mean square error (RMSE) and scale-invariant RMSE (si-RMSE) that measure the standard deviation of residuals of the rendered images. RMSE mainly evaluates the estimated lighting intensity, and si-RMSE focuses more on the evaluation of lighting positions. In addition, lighting estimation has also been evaluated by using Amazon Mechanical Turk (AMT) that performs crowdsourcing user study for subjective assessment of empirical realism of rendered images. 

Existing metrics evaluate estimated lighting largely by applying them to the rendered objects. Thus the performance of an estimation model is highly affected by test settings such as materials and 3D shape of the rendered objects. Based on the proposed spherical transport loss, we design a spherical transport distance (STD) metric that directly evaluates the optimal transport distance between the predicted illumination map and the ground-truth map on 
the unit sphere. The only difference from STL is that STD discards the auxiliary point strategy. The proposed STD highly simplifies the evaluation procedure and provides concise yet effective evaluations regarding to the lighting color, lighting intensity, and lighting position jointly.

\begin{figure}[ht]
\centering
\includegraphics[width=1.0\linewidth]{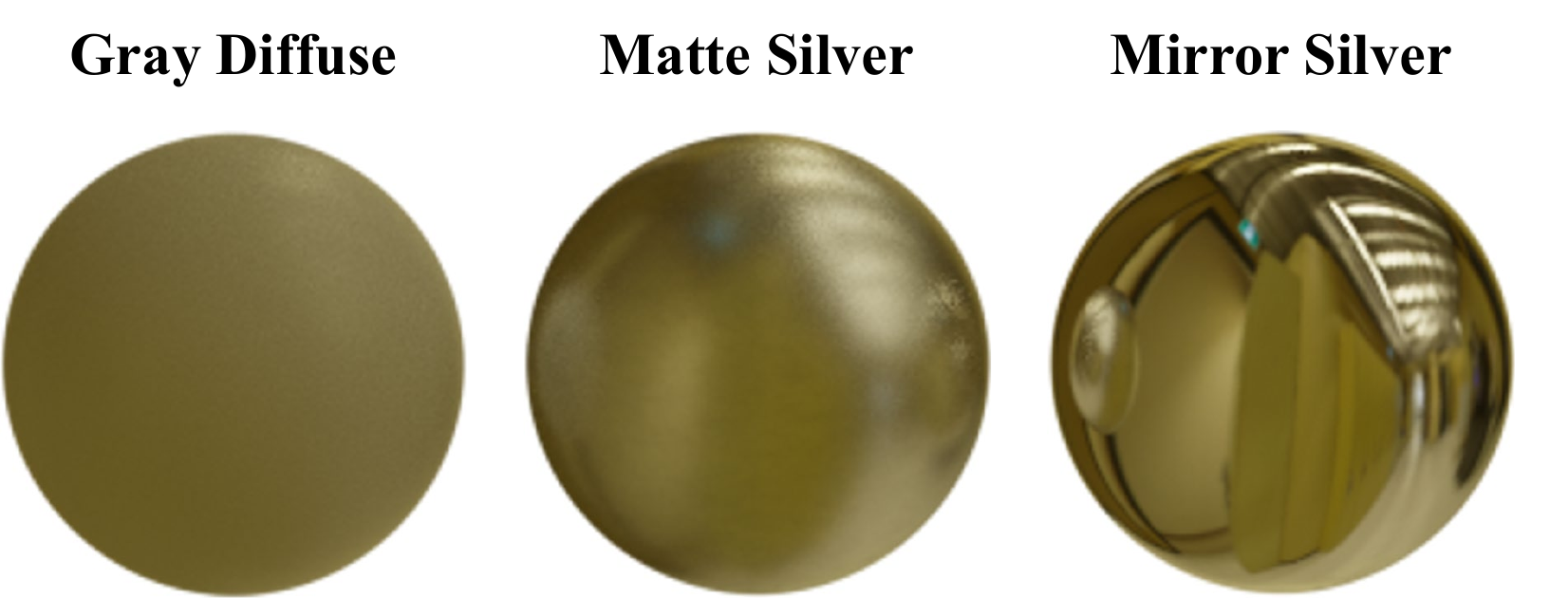}
\caption{The scene for quantitative evaluations consists of three
spheres with diffuse gray, matte silver and mirror silver materials.
}
\label{im_ball}
\end{figure}

\renewcommand\arraystretch{1.2}
\begin{table*}[ht]
\caption{
Comparison of NeedleLight with several state-of-the-art lighting estimation methods: The evaluation metrics include the widely used RMSE, si-RMSE, AMT, and our proposed STD. D, S, M denote a diffuse, a matte silver and a mirror material of the rendered objects, respectively.
}
\renewcommand\tabcolsep{3.8pt}
\centering 
\begin{tabular}{l|ccc|ccc|ccc|ccc|ccc} \hline
\multirow{1}{*} & 
\multicolumn{3}{c|}{\textbf{Gardner et al. \cite{gardner2017learning}}} & 
\multicolumn{3}{c|}{\textbf{Gardner et al. \cite{gardner2019deep}}} & 
\multicolumn{3}{c|}{\textbf{Li et al. \cite{li2019deep}}} &
\multicolumn{3}{c|}{\textbf{Garon et al. \cite{garon2019fast}}} &
\multicolumn{3}{c}{\textbf{NeedleLight}} \\
\cline{2-16}
\textbf{Metrics} & D & S & M & D & S & M& D & S & M & D & S & M & D & S & M \\\hline

\textbf{RMSE}    & 0.13 & 0.16 & 0.18   & \textbf{0.06} & 0.10 & 0.15   & 0.21 & 0.23 & 0.26   & 0.18 & 0.20 & 0.24    & 0.07 & \textbf{0.07} & \textbf{0.09}   \\

\textbf{si-RMSE} & 0.15 & 0.15 & 0.17   & 0.07 & 0.09 & 0.12   & 0.19 & 0.21 & 0.23   & 0.21 & 0.24 & 0.26    & \textbf{0.05} & \textbf{0.06} & \textbf{0.08}   \\

\textbf{AMT}     & 28\% & 23\% & 21\%   & 34\% & 33\% & 30\%   & 28\% & 27\% & 23\%   & 29\% & 26\% & 24\%    & \textbf{41\%} & \textbf{39\%} & \textbf{36\%}   \\\hline\hline


\textbf{STD}     &  & 6.84 &    &  & 5.52 &    &  & 7.01 &    &  & 7.14 &    &  & \textbf{4.21} &    \\\hline
\end{tabular}
\label{tab_compare}
\end{table*}

\section{Experiments}

\subsection{Dataset and Implementation}
We evaluate NeedleLight by using the Laval Indoor HDR Dataset \cite{gardner2017learning} that consists of 2,100 HDR panoramas taken in a variety of indoor environments. Similar to~\cite{gardner2017learning}, we extract eight limited field of view crops from each panorama which produces 19,556 images as used in our experiments. The image warping operation as described in \cite{gardner2017learning} is applied to the panoramas.
We apply the proposed sparse needlets with $j_{max} = 3$ to extract needlet coefficients as the ground truth for training.
Similar to \cite{gardner2019deep,garon2019fast}, DenseNet121 is used as the backbone network and produces a 4096-dimensional latent vector which is further forwarded to a fully-connected layer with 1024 units. Three separate output layers are added to regress needlet coefficients in frequency bands $j=1,2,3$.
In the experiment, we randomly select 200 images as testing set and the rest for training set. All the objects are rendered with Blender \cite{blender}.

The proposed NeedleLight is implemented by the PyTorch framework. The Adam is adopted as optimizer which employs a learning rate decay mechanism (initial learning rate is 0.001). The network is trained in 100 epochs with a batch size of 64. In addition, the network training is performed on two NVIDIA Tesla P100 GPUs with 16GB memory.

\subsection{Quantitative Evaluation}
We compare NeedleLight with a number of state-of-the-art lighting estimation methods including Garon et al. \cite{garon2019fast} that estimates lighting in frequency domain and Gardner et al. \cite{gardner2017learning,gardner2019deep} and Li et al. \cite{li2019deep} that estimate lighting in spatial domain. 
To perform quantitative evaluations, we create three spheres with gray diffuse, matte silver and mirror silver materials for evaluation as shown in Fig. \ref{im_ball}, which is consistent with the evaluation setting in \cite{legendre2019deeplight}. Then we render 300 images of objects (100 images for each material) by using the illumination maps that are predicted from testing set by each compared method. Table \ref{tab_compare} shows experimental results by using 4 evaluation metrics as described in \textit{Evaluation Metrics}, where \textit{D}, \textit{S} and \textit{M} denote a diffuse, a matte silver and a mirror material of the objects to be rendered, respectively.
The AMT user study is conducted by showing two images rendered by the ground truth and one of the methods in Table \ref{tab_compare} to 20 users who will pick the more realistic image. The score is the percentage of images rendered by the method that is deemed as more realistic than the ground-truth rendering.

We can observe that NeedleLight outperforms other methods in most cases under different evaluation metrics and materials as it allows regression in frequency and spatial domain jointly. 
The only exception is for diffuse material by \cite{gardner2019deep} while evaluated using RMSE, largely because the parameterization in Gardner et al. \cite{gardner2019deep} simplifies the scene illumination in spatial domain to achieve accurate prediction of light intensity while diffuse material is largely affected by light intensity and RMSE is most sensitive to light intensity.
Gardner et al. \cite{gardner2017learning} predicts illumination maps directly by a two-stage training strategy. As an under-constrained problem, the direct generation methods like Gardner et al. \cite{gardner2017learning} tend to over-fit the training set and present worse generalization performance.
Besides, both Gardner et al. \cite{gardner2017learning} and Gardner et al. \cite{gardner2019deep} estimate lighting in spatial domain which cannot recover frequency information and tends to generate inaccurate shading and shadow that are largely measured by si-RMSE.
Garon et al. \cite{garon2019fast} recovers lighting in frequency domain by regressing the SH coefficients, which tend to capture global instead of localized lighting. Thus Garon et al. \cite{garon2019fast} struggles to regress accurate lighting position and recover high frequency information.
Li et al. \cite{li2019deep} adopts spherical Gaussian functions to reconstruct illumination in spatial domain, thus it cannot recover accurate illumination frequency. Besides, it uses a masked L2 loss to preserve high frequency information though it cannot solve the missing of high frequency information essentially as illustrated in Fig. \ref{im_com}.
Instead, our proposed spherical transport loss with auxiliary points improves the regression of high frequency information significantly.

In addition, we can observe that the performance of the state-of-the-art methods is not consistent under different evaluation metrics. For example, Li et al. \cite{li2019deep} outperforms Garon et al. \cite{garon2019fast} in si-RMSE but the situation becomes the other way around in RMSE. The divergence of different metrics makes it hard to provide consistent evaluations. The proposed spherical transport distance (STD) instead provides relatively consistent and comprehensive evaluations in light intensity, color, position, etc. as shown in Table \ref{tab_compare}.

\begin{figure}[t]
\centering
\includegraphics[width=1.0\linewidth]{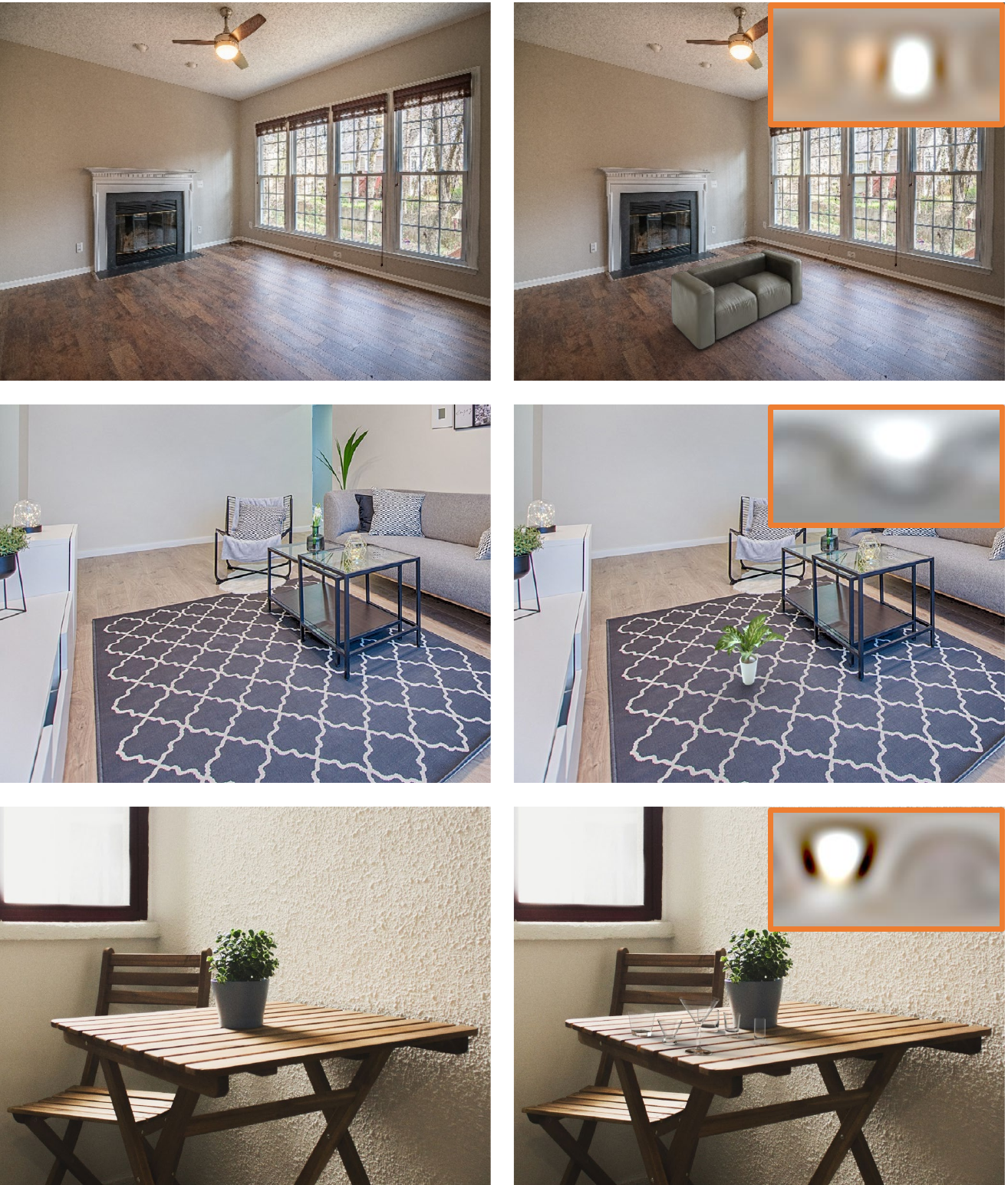}
\caption{
Object relighting on a variety of photos from the Internet. 
In all cases, light estimation is performed completely automatically by our model. 
The predicted illumination maps are utilized to relight the virtual objects with Blender \cite{blender}. 
More rendering samples and the analysis of \textbf{spatially-varying} rendering are available in supplementary file.
}
\label{im_real}
\end{figure}

\subsection{Qualitative Evaluation}

We also compare NeedleLight with four state-of-the-art lighting estimation methods qualitatively.
we design 25 3D scenes with objects for insertion and render them with the predicted illumination maps.
Fig. \ref{im_com} shows several rendered images and the predicted illumination maps. We can observe that NeedleLight predicts realistic illumination maps with plausible light sources, thus producing realistic rendering with clear and accurate shade and shadows that are very close to the ground truth. 
As a comparison, the illumination maps predicted by Gardner et al. \cite{gardner2017learning} and Gardner et al. \cite{gardner2019deep} tend to present clearly higher illumination frequency than the ground truth, largely because they recover illumination in spatial domain without explicitly taking frequency information into account.
Garon et al. \cite{garon2019fast} and Li et al. \cite{li2019deep} only predict illumination of low frequency, thus the produced renderings present very weak shade and shadow effects as illustrated in Fig. \ref{im_com}. 

Besides the testing set, we also validate the proposed method on natural images collected from the Internet as shown in Fig. \ref{im_real}. The proposed method achieves accurate estimation of scene illumination, thus the 3D objects can be embedded into the images with real shading and shadow effects.
We also include the analysis of \textbf{spatially-varying} illumination prediction in the supplementary file.

\renewcommand\arraystretch{1.2}
\begin{table*}[t]
\caption{Ablation study of the proposed NeedleLight: \textit{SH} denotes using spherical harmonics for representation; \textit{SN} denotes using spherical needlets for representation; \textit{ST} denotes applying the derived soft thresholding function to needlet coefficients; \textit{L2} and \textit{STL} denotes regressing coefficients with L2 loss and spherical transport loss;
\textit{SN+ST+L2+STL} denotes standard NeedleLight with sparse needlets which are regressed with both L2 and spherical transport loss.
}
\renewcommand\tabcolsep{8.25pt}
\centering 
\begin{tabular}{l||
ccc || 
ccc || 
ccc || 
p{1.25cm}<{\centering}} \hline
\multirow{2}{*} {\textbf{Models}} & 
\multicolumn{3}{c||}{\textbf{RMSE}} & 
\multicolumn{3}{c||}{\textbf{si-RMSE}} &
\multicolumn{3}{c||}{\textbf{AMT}} &
\multicolumn{1}{c}{\textbf{STD}} \\
\cline{2-11}
 & D & S & M & D & S & M& D & S & M & -   \\\hline

\textbf{SH+L2} & 0.19 & 0.20 & 0.25   & 0.22 & 0.24 & 0.28   & 23\% & 21\% & 18\%    & 7.14   \\


\textbf{SN + L2}     & 0.14 & 0.14 & 0.16   & 0.11 & 0.14 & 0.15   & 33\% & 31\% & 28\%    & 4.93   \\

\textbf{SN+ST+L2}     & 0.11 & 0.12 & 0.14   & 0.09 & 0.09 & 0.13   & 34\% & 32\% & 30\%    & 4.74    \\

\textbf{SN+ST+STL}     & 0.10 & 0.11 & 0.13   & 0.07 & 0.10 & 0.12   & 37\% & 34\% & 32\%    & 4.51   \\\hline

\textbf{SN+ST+L2+STL}         & \textbf{0.07} & \textbf{0.07} & \textbf{0.09}   & \textbf{0.05} & \textbf{0.06} & \textbf{0.08}   & \textbf{41\%} & \textbf{39\%} & \textbf{36\%}  & \textbf{4.21}   \\\hline
\end{tabular}
\label{tab_ablation}
\end{table*}

\renewcommand\arraystretch{1.1}
\begin{table}[t]
\caption{
Ablation studies over different bases including spherical harmonics (SH), spherical gaussian (SG), spherical distribution (SD), Haar, spherical needlets (SN). 
HT and ST denote apply hard thresholding and the derived soft thresholding functions to the spherical needlet coefficients.
$j_{max}$ denotes the order of needlets for representation. SN($j_{max}$=3)+ST is the standard setting of NeedleLight.
}
\renewcommand\tabcolsep{8.5pt}
\centering 
\begin{tabular}{l|p{1.5cm}<{\centering}|p{2cm}<{\centering}} \hline
\textbf{Models} & \textbf{L2} & \textbf{L2 + STL}
\\
\cline{2-3}
\hline 
\hline 

\textbf{SH}       & 7.14  & 7.13    \\
\textbf{SG}     & 6.01   &  5.74    \\
\textbf{SD}     & 5.75  &  5.54    \\
\textbf{Haar}      & 6.07  & 5.83    \\
\hline
\textbf{SN} ($j_{max}$=1)      & 7.52  & 7.18    \\
\textbf{SN} ($j_{max}$=2)      & 6.71  & 6.20   \\
\textbf{SN} ($j_{max}$=4)      & 5.37  & 4.96  \\
\hline

\textbf{SN}($j_{max}$=3)       & 5.32 & 4.93    \\
\textbf{SN}($j_{max}$=3)+\textbf{HT}    & 4.92 & 4.51    \\
\textbf{SN}($j_{max}$=3)+\textbf{ST}   & \textbf{4.74} & \textbf{4.21}    \\\hline

\end{tabular}
\label{tab_ablation2}
\end{table}

\subsection{Ablation Study}
We further evaluate NeedleLight by developing four NeedleLight variants as listed in Table \ref{tab_ablation}, including a baseline model which regress spherical harmonic coefficients with L2 loss (\textit{SH+L2}), regressing original needlet coefficients with L2 loss (\textit{SN+L2}), regressing needlet coefficients after applying soft thresholding (namely sparse needlets) with L2 loss (\textit{SN+ST+L2}), regressing sparse needlets coefficients with spherical transport loss (\textit{SN+ST+STL}). The standard NeedleLight regresses spares needlets with both L2 and spherical transport loss (\textit{SN+ST+L2+STL}).
Similar to the setting in Quantitative Evaluation, we apply the five variants and the standard NeedleLight to render 300 images with 15 objects of different materials. 
As Table \ref{tab_ablation} shows, 
using needlets for illumination representation (SN+L2) helps to achieve better lighting estimation compared with spherical harmonics function (SH+L2).
In addition, the performance of \textit{SN+ST+L2} is improved clearly as sparse needlet coefficients helps the regression of light sources significantly by the zero setting of redundant coefficients.
Additionally, the standard \textit{SN+ST+L2+STL} outperforms \textit{SN+ST+L2} and \textit{SN+ST+STL}, demonstrating the L2 and the proposed STL are complementary for the regression of needlet coefficients.

We also study effect of thresholding function and different orders of needlets and compare spherical needlets (SN) with other basis functions for illumination representation such as spherical harmonics (SH), spherical gaussian (SG), spherical distribution (SD) \cite{zhan2021emlight} and Haar \cite{mulcahy1997image} as shown in Table \ref{tab_ablation2}.
We followed the experimental setting as in Table \ref{tab_ablation} and measure the proposed STD as the evaluation metric. The number of coefficients of SH, SG, SD and Haar are set to be consistent with spherical needlets with 3 orders (about 250 coefficients).
As Table \ref{tab_ablation2} show, needlets SN($j_{max}$=3) outperforms other representation bases when regressing with both L2 loss and spherical transport loss.
Compared with spherical gaussian, spherical distribution and spherical harmonics functions, needlets enable representation in both spatial and frequency domains, thus achieving accurate regression in both domains.
Compared with other wavelets such as Haar, needlets is a new generation of spherical wavelets and is more suitable for the representation of spherical image with sharp local peak valleys which are commonly presented in HDR illumination map.
Regressing spherical harmonic coefficients with spherical transport loss doesn't improve the performance as there is no spatial information in spherical harmonic coefficients.
For spherical gaussian, spherical distribution and Haar, the performance with the spherical transport loss included is clearly improved as the corresponding coefficients are spatially localized, which demonstrates the effectiveness of the proposed spherical transport loss.
Further, the prediction performance drops when $j_{max}$=1,2 are used, and increasing the order to $j_{max}$=4 doesn't improve the performance. We conjecture that the larger number of parameters with $j_{max}$=4 affects the regression accuracy negatively.
Besides, the performance of SN($j_{max}$=3)+ST outperforms SN($j_{max}$=3)+HT, demonstrating the superiority of the derived soft thresholding function.

\section{Conclusions}
This paper presents NeedleLight, a lighting estimation model that introduces needlet basis for illumination representation and prediction. In NeedleLight, we deduce an optimal thresholding function from Bayesian framework for a sparse representation in terms of needlet basis. To tackle the regression of needlet coefficients with spatial localization, a novel spherical transport loss with auxiliary points is designed which performs regression by minimizing the discrepancy between two spherical distributions. Both the quantitative and qualitative experiments show that NeedleLight is capable of predicting illumination accurately from a single indoor image. We will continue to investigate needlet basis for more efficient illumination representation and explore optimal transport for better network training.

{\small
\bibliographystyle{ieee_fullname}
\bibliography{egbib}
}

\end{document}